\documentclass[10pt,conference]{IEEEtran}
\IEEEoverridecommandlockouts
\usepackage{cite}
\usepackage{amsmath,amssymb,amsfonts}
\newcommand{\clip}{\operatorname{clip}}

\usepackage[numbers,sort&compress]{natbib}
\usepackage{hyperref}
\hypersetup{
  colorlinks=false,
  pdfborder={0 0 0}
}
\usepackage{booktabs}
\usepackage{multirow}
\usepackage[section]{algorithm}
\usepackage{algorithmic}
\usepackage{algorithmic}
\usepackage{graphicx}
\usepackage{textcomp}
\usepackage{xcolor}
\def\BibTeX{{\rm B\kern-.05em{\sc i\kern-.025em b}\kern-.08em
    T\kern-.1667em\lower.7ex\hbox{E}\kern-.125emX}}
\begin{document}

\title{Heterogeneous Tasks Offloading in Vehicular Edge Computing: A Federated Meta Deep Reinforcement Learning Approach
}



\author{
\IEEEauthorblockN{
Yaorong Huang$^{*}$,
Jingtao Luo$^{\dagger}$,
Xuechao Wang$^{*}$
}

\IEEEauthorblockA{
$^{*}$The Hong Kong University of Science and Technology (Guangzhou)
}

\IEEEauthorblockA{
$^{\dagger}$Chengdu Neusoft University
}

\IEEEauthorblockA{
Email:
yhuang726@connect.hkust-gz.edu.cn,
LuoJingtao@nsu.edu.cn,
xuechaowang@hkust-gz.edu.cn
}
}

\maketitle

\begin{abstract}
Vehicular edge computing (VEC) enables latency-sensitive vehicular applications by offloading computation-intensive tasks to nearby edge servers. However, real-world vehicular workloads are typically modeled as heterogeneous directed acyclic graph (DAG) tasks with complex dependency structures, making joint offloading and resource allocation highly challenging. Moreover, distributed MEC deployment raises privacy concerns when collaboratively training learning-based policies.

In this paper, we propose a Federated Meta Deep Reinforcement Learning framework with GAT-Seq2Seq modeling (FedMAGS) for heterogeneous task offloading in VEC systems. The proposed approach leverages Graph Attention Networks to capture DAG dependencies, a Seq2Seq-based policy to generate structured offloading decisions, and federated meta-learning to enable fast adaptation across distributed MEC servers without sharing raw data.

Extensive simulations demonstrate that FedMAGS achieves faster convergence, lower execution delay, and better scalability compared with state-of-the-art baselines. In addition, the federated design preserves data privacy while reducing communication overhead, making the framework well suited for dynamic and large-scale VEC environments.
\end{abstract}


\section{Introduction}

The rapid development of vehicles, smart cities~\cite{chowdhury2025secure}, and intelligent transportation systems (ITS)~\cite{ahmad2024comprehensive} has significantly increased the demand for latency-sensitive vehicular applications, such as autonomous driving, traffic management, and large language model inference~\cite{musa2023sustainable, ma2024fedmg, 10591707}. These services require stringent real-time processing and high computational capability. However, vehicles are inherently constrained by limited onboard computing resources, making it difficult to satisfy such requirements locally.

Cloud computing was initially introduced to support vehicular task execution by offloading computation to remote data centers~\cite{dinh2013survey}. Although cloud-based solutions provide abundant computing resources, their centralized and geographically distant nature leads to substantial communication latency and instability under vehicle mobility~\cite{9373980}. Moreover, massive concurrent offloading requests may cause server-side resource contention and queuing delays~\cite{dong2024task}. These limitations motivate computation migration toward the network edge.

Mobile Edge Computing (MEC)~\cite{hu2015mobile} deploys computation resources closer to end devices, enabling low-latency and context-aware processing~\cite{mishra2023collaborative}. Vehicular Edge Computing (VEC), as a specialized MEC scenario, allows vehicles to offload tasks to nearby edge servers via wireless links. Compared with cloud-based solutions, VEC significantly reduces end-to-end delay and improves QoS. Nevertheless, efficient task offloading in practical VEC environments still faces several critical challenges.

\noindent\textbf{\textit{Resource Allocation.}}
Joint optimization of offloading decisions, computation scheduling, and resource allocation is generally NP-hard~\cite{guo2019toward}. Early studies adopted heuristic~\cite{liu2018offloading} and game-theoretic methods~\cite{guo2018collaborative, 10682861}, which lack adaptability in dynamic environments. Recently, deep reinforcement learning (DRL) has been applied to handle high-dimensional decision spaces~\cite{tang2020deep, 9212868}. However, many existing works assume simplified i.i.d. task models~\cite{10571777, 11023033}, limiting their generalization capability in heterogeneous scenarios.

\begin{figure}[htp!]
    \centering
    \includegraphics[width=1\linewidth]{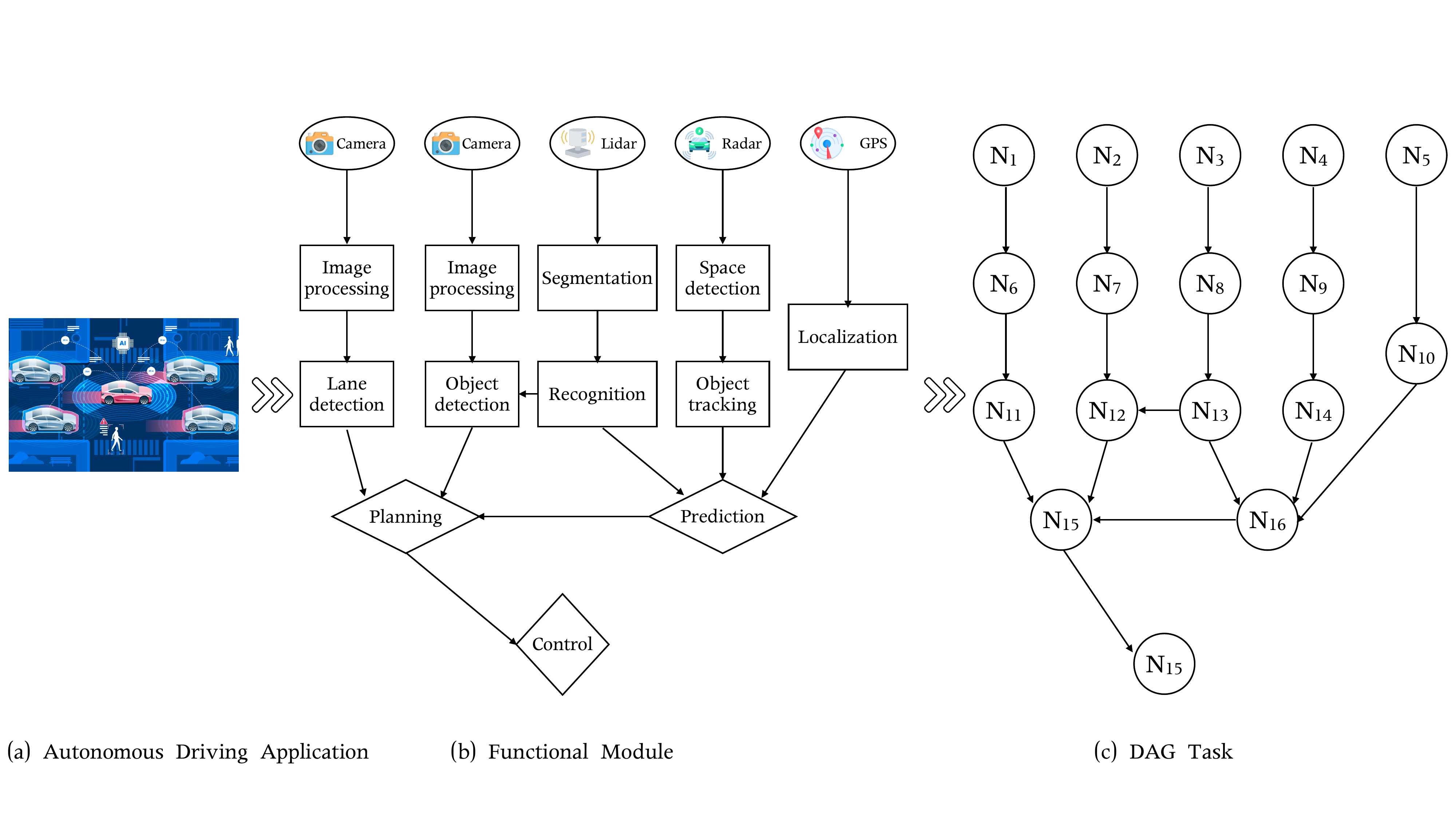}
    \caption{DAG Task of Autonomous Driving Application.}
    \label{fig:DAG_Tasks}
\end{figure}

\noindent\textbf{\textit{Heterogeneous DAG Tasks.}}
Real-world vehicular applications are commonly modeled as Directed Acyclic Graph (DAG) tasks, where subtasks exhibit complex dependency constraints and heterogeneous computation–communication characteristics. As illustrated in Fig.~\ref{fig:DAG_Tasks}, autonomous driving applications~\cite{sobhani2025modeling} consist of multiple interdependent modules with strict precedence constraints. Such structural heterogeneity significantly increases scheduling complexity, especially under limited edge resources and dynamic wireless conditions.

\noindent\textbf{\textit{Data Privacy.}}
In distributed VEC systems, each MEC server only observes local task data. Centralized aggregation of raw vehicular data for training raises privacy risks related to user behavior and location trajectories, while also incurring substantial communication overhead. Therefore, privacy-preserving collaborative learning without raw data sharing is essential.

\noindent\textbf{\textit{This Work.}}
To address the above challenges, we propose a Federated Meta Deep Reinforcement Learning framework with Graph Attention Network and Seq2Seq modeling (FedMAGS). The proposed approach integrates:
(i) GAT-based dependency-aware task representation,
(ii) Seq2Seq-based structured offloading decision generation, and
(iii) federated meta-learning for rapid adaptation across distributed MEC servers without sharing raw data.

This unified design enables adaptive, scalable, and privacy-preserving task offloading for heterogeneous DAG workloads in VEC networks.

\noindent\textbf{\textit{Contributions.}} The main contributions are summarized as follows:
\begin{itemize}
    \item \textbf{Problem Formulation:} We formulate heterogeneous DAG-based task offloading in VEC as a joint optimization problem integrating dependency constraints, resource allocation, and privacy considerations.
    \item \textbf{FedMADRL Framework:} We design a federated meta-DRL framework that enables collaborative and adaptive policy learning across MEC servers without raw data exchange.
    \item \textbf{Dependency-Aware Modeling:} We incorporate a GAT-Seq2Seq architecture to capture DAG topology and critical-path information for structured offloading decisions.
    \item \textbf{Extensive Evaluation:} Simulation results demonstrate superior convergence, scalability, and latency performance compared with state-of-the-art baselines.
\end{itemize}

\section{Related Work}

\subsection{DRL-based Task Offloading}
In highly dynamic MEC environments, traditional heuristic and game-theoretic methods often struggle to cope with high-dimensional state spaces and time-varying network conditions. Consequently, deep reinforcement learning (DRL) has emerged as a dominant approach for adaptive task offloading and resource allocation.
Wu et al.~\cite{wu2025qoe} proposed QoE-aware DRL algorithms based on the Proximal Policy Optimization (PPO) framework to optimize resource allocation for AIGC services supported by MEC.
To address economic interactions in vehicular networks, Zhao et al.~\cite{zhao2025game} incorporated a TD3-based DRL algorithm into a Stackelberg game framework to jointly optimize RSU channel access and dynamic pricing. Li et al.~\cite{li2025collaborative} developed a distributed task offloading and resource allocation scheme based on multi-agent PPO (MAPPO) to minimize user energy consumption under delay constraints in small-cell MEC networks.
Furthermore, integrating digital twin technology with DRL has recently attracted attention for enhancing mobility awareness. For example, Chen et al.~\cite{chen2025mobility} leveraged digital twin–assisted DRL to predict future user states and optimize QoS-aware task offloading decisions.
Although these DRL-based approaches improve adaptability compared to conventional optimization methods, most of them assume simplified task structures and rely on centralized or fixed-policy learning, limiting their ability to generalize across heterogeneous DAG-based workloads and distributed privacy-constrained environments.

\subsection{DAG-Based Task Modeling and Scheduling}
To capture the intrinsic dependency structures of real-world applications, modern vehicular workloads are increasingly modeled as Directed Acyclic Graph (DAG) tasks. Recent research has focused on dependency-aware scheduling to reduce end-to-end latency.
Deng et al.~\cite{deng2025task} proposed a DRL-based task offloading scheme (DVTP) that integrates Variational Graph Attention Networks and Transformer models for DAG task scheduling in vehicular networks. Zhao et al.~\cite{Zhao2025dynamic} introduced a cache and dependency-aware task offloading framework (CachOf) that combines priority computation and DRL to enhance application execution efficiency. Dai et al.~\cite{Dai2024meta} developed a Seq2Seq-based meta reinforcement learning algorithm (SMRL-MTO) for VEC task offloading, which explores DAG subtask dependencies via Seq2Seq model and optimizes task execution time.
While these studies explicitly consider DAG structures, they typically focus on centralized learning or standalone dependency modeling, without integrating distributed privacy-preserving training mechanisms in large-scale MEC systems.

\subsection{Federated Learning for Task Offloading}
Federated learning (FL) has emerged as a promising paradigm for privacy-preserving collaborative learning in distributed MEC environments. By enabling multiple edge servers to jointly train models without sharing raw data, FL mitigates privacy risks and reduces communication overhead.
Chen et al.~\cite{chen2024pfr} proposed a personalized federated DRL-based offloading and resource allocation scheme (PFR-OA) to adapt to dynamic system states and heterogeneous user demands in multi-edge intelligent communities. Similarly, Chen et al.~\cite{chen2025mcmfdrl} introduced a multi-agent collaboration method (MCM-FDRL) based on federated deep reinforcement learning to enhance task offloading efficiency in large-scale vehicular MEC scenarios.
Despite these advancements, existing FL-based approaches rarely consider heterogeneous DAG-based task structures or incorporate meta-learning mechanisms to enable rapid generalization across diverse vehicular workloads.

\subsection{Summary and Research Gap}
In summary, prior works have investigated DRL-based adaptive resource allocation, DAG-aware task scheduling, and federated learning for privacy-preserving optimization in MEC systems. However, these research directions are largely explored in isolation. Limited efforts have jointly addressed heterogeneous DAG task modeling, meta deep reinforcement learning for rapid adaptation, and federated privacy-preserving training in distributed vehicular MEC environments.

This gap motivates the Federated Meta Deep Reinforcement Learning framework with GAT-Seq2Seq modeling proposed in this work.

\section{System Architecture}
\label{sec:system_model}
\begin{figure*}[htp!]
\centering
\includegraphics[width=0.75\textwidth]{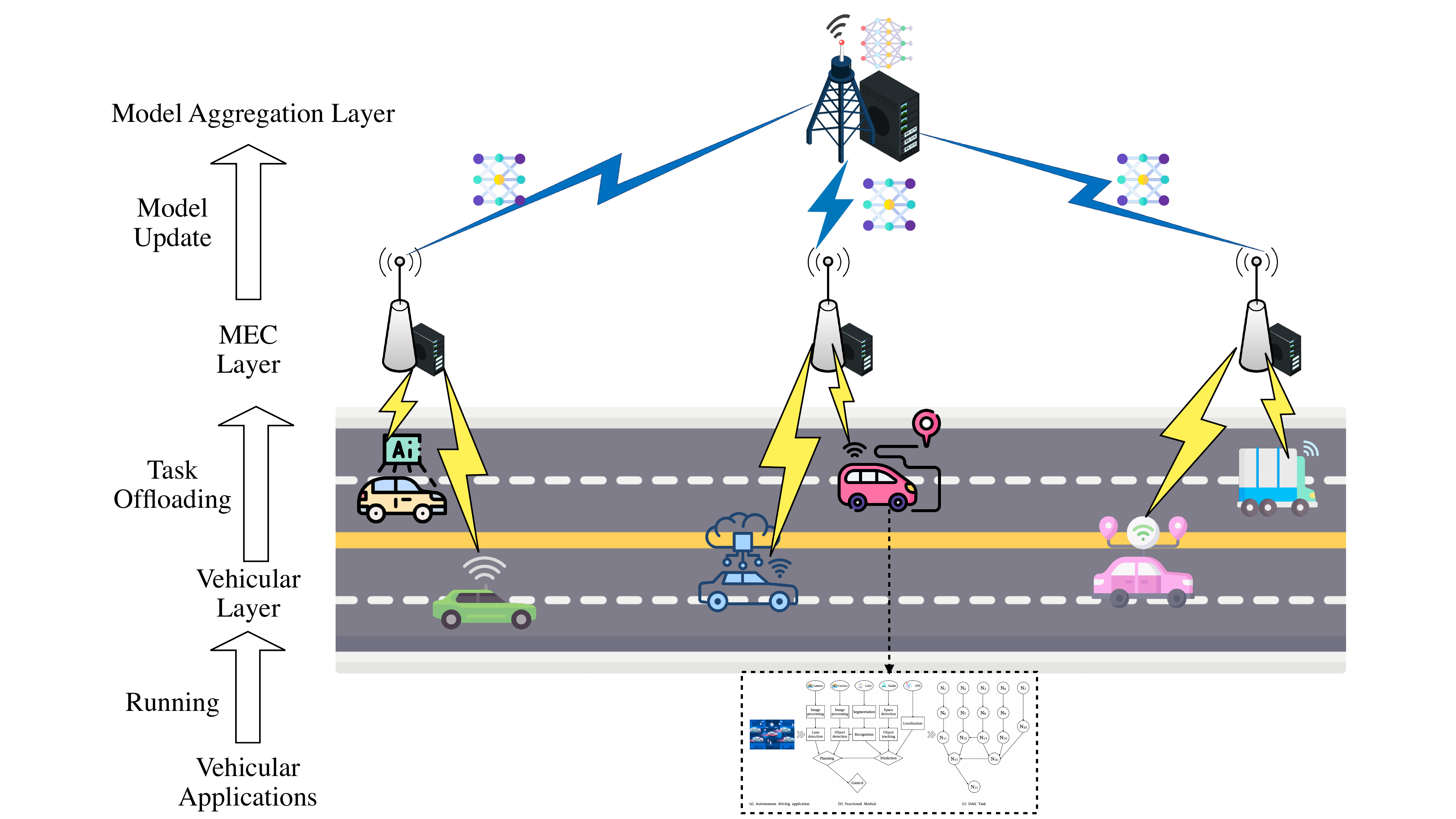}
\caption{The System Architecture Heterogeneous Task offloading in VEC networks}%
\label{fig:architecture}
\end{figure*}

Fig.~\ref{fig:architecture} illustrates the proposed distributed heterogeneous task offloading architecture in VEC networks. The system consists of three hierarchical layers: the vehicular application layer, the vehicular layer, and the MEC layer with a federated aggregation server.




\subsection{Vehicular Application Layer}

At the lowest layer of the architecture, vehicles run a variety of intelligent applications, such as autonomous driving service. These applications are both computation-intensive and subject to stringent end-to-end latency requirements.

Each application is decomposed into multiple interdependent functional modules, which are collectively represented as a DAG. In the DAG representation, nodes represent individual subtasks, while directed edges indicate both data dependencies and strict execution precedence constraints. Such DAG formulation naturally captures the heterogeneous nature of modern vehicular workloads, where subtasks exhibit substantial variations in computational complexity, input/output data sizes, and individual latency tolerances.

For clarity, we adopt the following terminology: a subtask is referred to as a \textit{parent} of another if there exists a directed edge from the former to the latter; conversely, the receiving subtask is termed its \textit{child}. A child subtask may commence execution only after it has received (i) the computation results from \textit{all} of its parent subtasks and (ii) any auxiliary metadata specific to the subtask itself.

Each vehicular task is thus fully characterized by its DAG topology, the computation demand of each subtask, and the data volume transferred along each dependency edge. These parameters vary significantly across different application types.
In the proposed framework, the subtask constitutes the atomic unit of offloading: each subtask is indivisibly assigned to either local onboard execution or remote execution at an MEC server.


\subsection{Vehicular Layer}

In the vehicular layer, each vehicle generates multiple computation tasks over time, which can be partially or fully processed using its onboard computation resources. Due to the limited computing capability of onboard processors, vehicles with complex task topologies or large numbers of subtasks must offload a subset of subtasks to nearby MEC servers.

To facilitate fine-grained subtask management and scheduling, each vehicle maintains two separate queues: a local computation queue for subtasks awaiting execution on the onboard processor, and a local transmission (or upload) queue for subtasks to be offloaded to MEC servers via wireless links. Computation and transmission operations can proceed concurrently, enabling pipelined processing and improved overall task execution efficiency.

Prior to offloading, vehicles periodically encapsulate and report subtask metadata—including required computation cycles, input data size, output data size, and dependency information—to the associated MEC servers via wireless channels. Leveraging this information, MEC servers perform coordinated subtask placement and resource allocation to minimize offloading-induced latency while respecting instantaneous channel conditions and server-side resource availability.

This collaborative offloading paradigm between vehicles and MEC servers supports flexible adaptation to diverse application requirements and heterogeneous resource constraints, thereby enhancing the responsiveness and scalability of vehicular edge computing systems.

\subsection{MEC Layer}

Multiple MEC servers are deployed along roadside infrastructures to provide edge computing services for vehicles within their coverage areas. Each MEC server is equipped with limited computational resources and wireless communication channels to support concurrent subtask execution and data transmission.

Upon receiving task descriptions from associated vehicles, the MEC server is responsible for coordinating subtask execution locations and allocating communication and computation resources. Due to the DAG-based structure of vehicular applications, the execution of each subtask is constrained by precedence dependencies and data transmission requirements.

\begin{figure}[t!]
\centering
\includegraphics[width=0.5\textwidth]{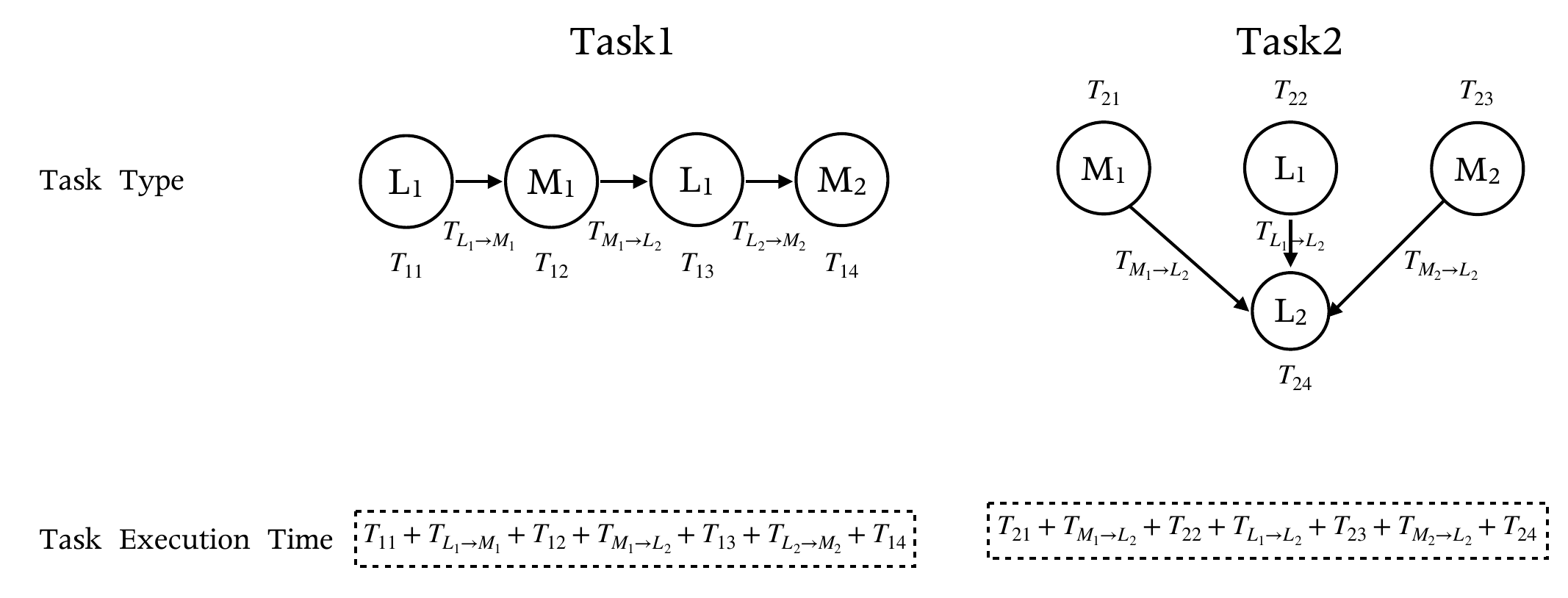}
\caption{The Examples of DAG Task Modeling}%
\label{fig:task_example}
\end{figure}

As illustrated in Fig.~\ref{fig:task_example}, subtasks can be executed either locally at the vehicle (denoted by $L$) or at the MEC server (denoted by $M$). When two dependent subtasks are executed at different locations, additional transmission latency is incurred. Therefore, the overall task completion time consists of local computation delay, edge computation delay, and cross-location data transmission delay along the dependency chain.

For example, in a sequential DAG (Task 1), the task execution time accumulates along the entire dependency path. In contrast, for a branched DAG (Task 2), multiple subtasks may execute in parallel, while the final completion time is determined by the critical path involving both computation and transmission delays. Consequently, improper offloading decisions may introduce cascading latency propagation across dependent subtasks.

Due to dynamic vehicle mobility and time-varying wireless conditions, the workload distribution and channel quality observed by each MEC server change over time. To adapt to such dynamics, each MEC server locally trains a meta deep reinforcement learning agent through continuous interaction with its environment. The agent learns to make dependency-aware task offloading and resource allocation decisions based on locally observed task states, resource availability, and communication conditions.

Importantly, all training procedures are conducted locally at each MEC server without sharing raw vehicular task data, ensuring privacy preservation and low-latency edge operation. The locally updated models are subsequently synchronized via the upper-layer model aggregation mechanism.

\subsection{Model Aggregation Layer}

To enable collaborative learning across geographically distributed MEC servers while preserving data privacy, a model aggregation layer is introduced at the top of the architecture.

In the proposed federated framework, each MEC server independently trains its local task offloading and resource allocation policy using locally observed vehicular task data. Due to the distributed deployment and heterogeneous task distributions across different regions, the locally trained models may exhibit performance variations.

Instead of transmitting raw vehicular data, which may contain sensitive information such as location traces and driving behaviors, each MEC server periodically uploads only its model parameters to the aggregation server through backhaul links. The aggregation server performs parameter fusion to generate a global model that captures shared knowledge across multiple MEC regions.

After aggregation, the updated global model is redistributed to all MEC servers to initialize the next round of local training. Through iterative local updates and global aggregation, the system achieves collaborative policy learning without exposing raw task data.

This model aggregation layer enhances scalability and generalization capability across heterogeneous vehicular environments, while effectively preserving data privacy and reducing communication overhead.

\section{Problem Formulation} \label{sec:problem_formulation}

\subsection{Preliminaries}

We consider a set of task offloading in VEC scenarios $\{\mathcal{T}_i\}$, where each scenario $\mathcal{T}_i$ is characterized by the vehicular tasks and the MEC server parameters. In scenario $\mathcal{T}_i$, let $V$ denote the set of vehicles. Each vehicle $v \in V$ generates one computation task modeled as a DAG. The DAG consists of a set of subtasks and directed edges representing precedence and data dependencies.

The $p$-th subtask of vehicle $v$ is denoted by $t_{v,p}$ and characterized by the triplet $(d_{v,p}^L, c_{v,p}, d_{v,p}^O)$, where
\begin{itemize}
    \item $d_{v,p}^L$ is the input size of the subtask itself (program code and parameters),
    \item $c_{v,p}$ is the required computation workload (in CPU cycles),
    \item $d_{v,p}^O$ is the output data size after execution.
\end{itemize}

The total input data size of subtask $t_{v,p}$, denoted $d_{v,p}^I$, is the sum of its own input size and the output sizes of all its parent subtasks:
\begin{equation} \label{eq:input_data}
d_{v,p}^I = d_{v,p}^L + \sum_{t_{v,k} \in P(t_{v,p})} d_{v,k}^O, \quad \forall v \in V, \ \forall p \in P,
\end{equation}
where $P(t_{v,p})$ is the set of parent subtasks of $t_{v,p}$. We assume that sibling subtasks (sharing the same parents) receive identical copies of the parents' output data.

The MEC server is equipped with $R$ orthogonal uplink subchannels, where the $r$-th subchannel has bandwidth $w_r$ ($r=1,\dots,R$), and $M$ computation processors, where the $m$-th processor operates at frequency $f_m$ ($m=1,\dots,M$). For notational convenience, we introduce index $0$ to represent local vehicle execution.

The offloading decision for subtask $t_{v,p}$ is jointly determined by binary variables $x_{v,p}^r \in \{0,1\}$ ($r=0,1,\dots,R$) and $y_{v,p}^m \in \{0,1\}$ ($m=0,1,\dots,M$), where $x_{v,p}^r = 1$ ($y_{v,p}^m = 1$) indicates that the subtask is offloaded using the $r$-th uplink subchannel (the $m$-th processor). Local execution is indicated by $x_{v,p}^0 = 1$ or equivalently $y_{v,p}^0 = 1$.

Since each subtask is indivisible and assigned to exactly one uplink channel and one processor (or local resources), the following constraints hold:
\begin{subequations} \label{eq:offloading_constraints}
\begin{align}
    \sum_{r=0}^{R} x_{v,p}^r &= 1, \quad \forall v \in V, \ \forall p \in P, \label{eq:channel_choice} \\
    \sum_{m=0}^{M} y_{v,p}^m &= 1, \quad \forall v \in V, \ \forall p \in P, \label{eq:processor_choice} \\
    x_{v,p}^0 &= y_{v,p}^0, \quad \forall v \in V, \ \forall p \in P, \label{eq:local_consistency} \\
    x_{v,p}^r, \ y_{v,p}^m &\in \{0,1\}, \quad \forall r,m.
\end{align}
\end{subequations}

Both computation and transmission queues at vehicles and the MEC server are modeled as first-in-first-out (FIFO) queues.

\subsection{Task Offloading and Delay Model} \label{subsec:task_offloading_model}

The end-to-end execution delay of a task comprises transmission delay (upload/download) and computation delay, accounting for queueing and precedence constraints.

\subsubsection{Input Data Transmission Delay}

Let $TE(t_{v,p})$ denote the time instant when all input data of subtask $t_{v,p}$ has been received at the execution location. We distinguish two cases based on the offloading decision.

\textbf{Case 1: Local execution} ($x_{v,p}^0 = 1$ or $y_{v,p}^0 = 1$).
Only the outputs of parents executed at the MEC server, i.e., $\{d_{v,k}^O \mid t_{v,k} \in P^M(t_{v,p})\}$, need to be downloaded to the vehicle. Upon completion of parent $t_{v,k}$ at time $CE(t_{v,k})$, its output is enqueued in the downlink queue. The transmission start time is
\begin{equation}
TS(d_{v,k}^O) = CE(t_{v,k}) + WT^d(t_{v,k}),
\end{equation}
where $WT^d(t_{v,k})$ is the waiting time in the downlink queue. According to the Shannon's formula, the transmission duration over the downlink channel is
\begin{equation}
T(d_{v,k}^O) = \frac{d_{v,k}^O}{w_d \log_2 \left(1 + \frac{q_m g_v}{\varpi}\right)},
\end{equation}
with $w_d$ the downlink bandwidth, $q_m$ the MEC transmit power, $g_v$ the channel gain from the MEC server to vehicle $v$, and $\varpi$ the noise power.

Thus,
\begin{equation}
TE(t_{v,p}) = \max_{t_{v,k} \in P^M(t_{v,p})} \Bigl\{ TS(d_{v,k}^O) + T(d_{v,k}^O) \Bigr\}.
\end{equation}

\textbf{Case 2: MEC offloading} ($x_{v,p}^r = 1$ for some $r \geq 1$).
Both the subtask itself ($d_{v,p}^L$) and the outputs of locally executed parents $\{d_{v,k}^O \mid t_{v,k} \in P^L(t_{v,p})\}$ must be uploaded. The vehicle aggregates these data into a single upload burst. The latest-completed local parent is
\begin{equation}
t^*_{v,p} = \arg\max_{t_{v,k} \in P^L(t_{v,p})} CE(t_{v,k}).
\end{equation}
Its upload start time is
\begin{equation}
TS(t^*_{v,p}) = CE(t^*_{v,p}) + WT^r(t^*_{v,p}),
\end{equation}
where $WT^r(t^*_{v,p})$ is the waiting time in the $r$-th uplink queue. The upload duration is
\begin{equation}
T(t^*_{v,p}) = \frac{d_{v,p}^L + \sum_{t_{v,k} \in P^L(t_{v,p})} d_{v,k}^O}{w_r \log_2 \left(1 + \frac{q_v g_v}{\varpi}\right)},
\end{equation}
with $q_v$ the vehicle transmit power. Hence,
\begin{equation}
TE(t_{v,p}) = TS(t^*_{v,p}) + T(t^*_{v,p}).
\end{equation}

\subsubsection{Computation Delay}

After all input data arrives, subtask $t_{v,p}$ is enqueued at the assigned computation resource. The computation start time is
\begin{equation}
CS(t_{v,p}) = TE(t_{v,p}) + WT^\ell(t_{v,p}),
\end{equation}
where $WT^\ell(t_{v,p})$ is the queueing delay at the local processor ($\ell = v$) or MEC processor ($\ell = m$).

The execution duration is
\begin{equation}
C(t_{v,p}) =
\begin{cases}
\frac{c_{v,p}}{f_v} & \text{if } y_{v,p}^0 = 1, \\
\frac{c_{v,p}}{f_m} & \text{if } y_{v,p}^m = 1.
\end{cases}
\end{equation}

The completion time of the subtask is therefore
\begin{equation}
CE(t_{v,p}) = CS(t_{v,p}) + C(t_{v,p}).
\end{equation}

The end-to-end execution time of the entire task on vehicle $v$ (from the earliest submission time $ST(t_v)$ to task completion) is determined by the critical path:
\begin{equation} \label{eq:task_execution_time}
ET(t_v) = \max_{p} CE(t_{v,p}) - ST(t_v).
\end{equation}

The system-level performance metric is the average execution time (AET) over all vehicles:
\begin{equation} \label{eq:aet}
AET = \frac{1}{|V|} \sum_{v \in V} ET(t_v).
\end{equation}

\subsection{Optimization Problem}

Each scenario $\mathcal{T}_i$ is defined by a specific realization of task DAGs, channel bandwidths $\{w_r, w_d\}$, uplink/downlink channel gains $\{g_v\}$, and computation capacities $\{f_v, f_m\}$, which are assumed to be drawn from given distributions reflecting real-world dynamics.

The joint subtask offloading and resource allocation problem in scenario $\mathcal{T}_i$ is formulated as
\begin{equation} \label{eq:optimization_problem}
\begin{aligned}
& \underset{\{x_{v,p}^r\}, \{y_{v,p}^m\}}{\min} \quad AET \\
& \text{s.t.} \quad \text{Constraints~\eqref{eq:offloading_constraints}}.
\end{aligned}
\end{equation}

Problem~\eqref{eq:optimization_problem} is a mixed-integer nonlinear program (MINLP) due to the max and argmax operators in delay expressions and the coupling between offloading decisions and queueing/transmission/computation delays. Conventional optimization-based or heuristic approaches incur prohibitively high computational overhead and lack fast adaptability to highly dynamic vehicular environments, motivating the design of learning-based methods that can generalize across heterogeneous scenarios.

\section{Algorithm}
\label{sec:algorithm_design}
\subsection{MDP Formulation with GAT-Seq2Seq Encoding}

We model heterogeneous task offloading in dynamic VEC networks as a family of Markov Decision Processes (MDPs), where each offloading scenario
$\mathcal{T}_i \sim \rho(\mathcal{T})$ corresponds to one MDP.

For scenario $\mathcal{T}_i$, the MDP is defined by the tuple
\[
\mathcal{M}_i = (\mathcal{S}, \mathcal{A}, \mathcal{R}, \gamma),
\]
where $\mathcal{S}$, $\mathcal{A}$, $\mathcal{R}$ and $\gamma$ denote the state space, the action space, the reward function, and the discount factor, respectively.

\subsubsection{State Space}
The state $s_{i}$ of the subtask $t_{i}$ is defined as

\begin{equation}
\begin{aligned}
s_{i} &= \left( \mathbf{F}_{i}, I^p_{i}, I^s_{i},\right), \\
\mathcal{S} &= \{ s_{i} \mid i = 1,2,\dots,N \},
\end{aligned}
\label{eq:state_refined}
\end{equation}

where:

\begin{itemize}
\item $\mathbf{F}_{i} = [I_{i}, \, d_{i}^L, \, c_{i}, \, f_i, \, \{f_m\}_{m=1}^M, \, \{w_r\}_{r=1}^R]$ is the feature vector of subtask $t_{i}$, where $I_{i}$ is the topological index of the subtask in the DAG, $f_i$ is the local CPU frequency of vehicle $i$, and $\{f_m\}$, $\{w_r\}$ are the frequencies and bandwidths of MEC resources.
\item $I^p_{i}$ is the parent's node position of subtask $t_{i}$;
\item $I^s_{i}$ is the son's node position of subtask $t_{i}$.
\end{itemize}
To encode the variable DAG structure while preserving dependencies and enabling sequential decision generation, we employ a Graph Attention Network (GAT)~\cite{velivckovic2017graph} followed by a Seq2Seq model.

\subsubsection{Topological Feature Extraction via GAT}
To capture the complex dependencies in the DAG structure of subtasks, we utilize a GAT to extract topological features.

For subtask $t_{i}$, the GAT aggregates high-dimensional feature representation $F'_i$ via multi-head attention by feature
$F_i$ and the adjacency matrix $\mathcal{G}$ of the DAG:
\begin{equation} \label{eq:gat_embedding}
F'_i = \Big\| _{k=1}^K \sigma \left( \sum_{j \in \mathcal{N}_i} \text{atten}_{ij}^k \mathbf{W}^k F_j \right),
\end{equation}
where $\text{atten}_{ij}^k$ represents attention coefficients.
The GAT embeddings $F'_i$ are concatenated with dependency vectors $I_i^p$ and $I_i^s$ of subtask $t_i$, forming input $H_i$:
\begin{equation} \label{eq:aggregated_input}
H_i = [F'_i, I_i^p, I_i^s].
\end{equation}

\subsubsection{Seq2Seq Decision Generation} The Seq2Seq network maps the non-Euclidean DAG structure into a serialized offloading plan, effectively reducing the action space complexity from exponential to linear in the number of subtasks.

The aggregated features $\{H_i\}$ are ordered according to a topological sort of the DAG and fed into a bidirectional gated recurrent unit (BiGRU) encoder, which captures both forward and backward dependencies. The encoder's hidden states are then processed by an attentional GRU decoder that generates sequential offloading decisions for each subtask, allowing the policy to consider the entire DAG context when making decisions.

The encoder computes hidden states $e_i$:
\begin{equation} \label{eq:lstm_encoder}
\begin{split}
&e_i^{f} = h^{f}_{enc}(e_{i-1}^{f}, H_i),\\
&e_i^{b} = h^{b}_{enc}(e_{i+1}^{b}, H_{i+1}),\\
&e_i = (e_i^{f}, e_i^{b}).
\end{split}
\end{equation}

The decoder output $d_j$ is:
\begin{equation} \label{eq:lstm_decoder}
d_j = h_{dec}(d_{j-1},c_j,a_{j-1}),
\end{equation}
where $c_j$ is the context vector.

And the final offloading decision sequence of the DAG task $A_{DAG}$ is obtained by the decoder output ${d_j}$ via softmax layer:
\begin{equation} \label{eq:offloading_sequence}
A_{DAG} = \text{Softmax}(d).
\end{equation}

\subsubsection{Action Space}
The action $a_t$ at time step $t$ corresponds to the offloading decision for subtask $t_i$, which is a categorical variable indicating the choice of local execution or one of the MEC resources (uplink channels and processors). The action space $\mathcal{A}$ is defined as
\begin{equation}
A_{DAG} =\{a_1, a_2, \dots, a_N\} \in \mathcal{A},
\end{equation}
\begin{equation}
\mathcal{A} = \{0\} \cup \{(r,m) \mid r=1,\dots,R; m=1,\dots,M\},
\end{equation}
where action $0$ denotes local execution, and action $(r,m)$ denotes offloading using the $r$-th uplink subchannel and the $m$-th processor at the MEC server.

\subsubsection{Reward Function}

To align step-wise optimization with minimizing total execution time,
we define incremental delay reward:

\begin{equation} \label{eq:reward_design}
  r_t =
  \begin{cases}
  CE(t_{j}) - CE(t_{i}) & \text{if subtask } t_{j} \text{ follows } t_{i}, \\
  -CE(t_{i}) & \text{if } t_{i} \text{ is the first subtask}.
  \end{cases}
\end{equation}
The cumulative return is

\begin{equation}
G = \sum_{t=1}^{T} \gamma^{t-1} r_t.
\end{equation}

 Higher rewards reflect reduced completion time increments, aligning with minimizing AET in Section~\ref{sec:problem_formulation}.

\subsection{Federated Meta Policy Learning}

We adopt a model-agnostic meta-learning (MAML)~\cite{finn2017model} backbone integrated with federated learning aggregation~\cite{mcmahan2017communication} to enable privacy-preserving collaborative training across distributed MEC servers, each observing heterogeneous local task distributions.

Each MEC server $k$ maintains a local policy network $\pi_{\theta_k}$ (parameterized by GAT encoder + actor-critic heads) and value network $V_{\phi_k}$. Training alternates between:

1. \textbf{Local Adaptation (Inner Loop)}: Using Proximal Policy Optimization (PPO)~\cite{schulman2017proximal} on local trajectories sampled from $\mathcal{T}_i \sim \rho_k(\mathcal{T})$ (region-specific distribution). The clipped surrogate objective is
\begin{equation}\label{eq:ppo_clip}
\begin{aligned}
L^{\text{CLIP}}(\theta)
= \hat{\mathbb{E}}_t \Big[
\min\big(& \rho_t(\theta) \hat{A}_t, \\
& \clip(\rho_t(\theta), 1-\epsilon, 1+\epsilon)\hat{A}_t
\big) \Big]  \\
& - \beta \mathrm{KL}[\theta' \| \theta]
\end{aligned}
\end{equation}

with value loss $L^V(\phi) = \hat{\mathbb{E}}_t [(V_\phi(s_t) - \hat{R}_t)^2]$, where $\rho_t(\theta) = \pi_\theta(a_t|s_t)/\pi_{\theta'}(a_t|s_t)$ is the importance ratio, $\hat{A}_t$ is GAE, and $\hat{R}_t$ is discounted return. After $m$ inner steps, adapted parameters are $\hat{\theta}_k = \theta_k + \alpha \nabla_\theta L(\theta_k)$.

2. \textbf{Federated Meta Update (Outer Loop)}: Each MEC uploads only model deltas $(\hat{\theta}_k - \theta_k)$ to the aggregation server. The server performs FedAvg-style weighted averaging to obtain global meta-parameters:
\begin{equation} \label{eq:fed_meta_update}
\theta \leftarrow \theta + \beta \frac{1}{K} \sum_{k=1}^K (\hat{\theta}_k - \theta),
\end{equation}
approximating first-order MAML to avoid expensive second-order gradients while enabling fast adaptation (few local PPO steps) to unseen $\mathcal{T}_i$.

This federated meta-DRL paradigm ensures: (i) raw task/DAG data never leaves vehicles/MEC coverage, (ii) meta-knowledge aggregates heterogeneous regional patterns, and (iii) GAT captures dependency-induced heterogeneity critical for DAG offloading.

The complete procedure is outlined in Algorithm~\ref{alg:smrl_fed_gat}.

\begin{algorithm}[t!]
\caption{Federated Meta PPO with GAT-Seq2Seq for Heterogeneous Task Offloading (FedMAGS)}
\label{alg:smrl_fed_gat}
\renewcommand{\algorithmicrequire}{\textbf{Input}}
\renewcommand{\algorithmicensure}{\textbf{Output}}
\begin{algorithmic}[1]
\small
\REQUIRE Task distribution $\rho(\mathcal{T})$; $K$ MEC servers; meta-params $\theta$; inner steps $m$; outer rounds $K_{\text{outer}}$; adaptation lr $\alpha$, meta lr $\beta$
\FOR{outer round $=1$ to $K_{\text{outer}}$}
    \FOR{each MEC server $k=1$ to $K$ \textbf{in parallel}}
        \STATE Sample batch of scenarios $\{\mathcal{T}_i\}_{i=1}^B \sim \rho_k(\mathcal{T})$
        \FOR{each $\mathcal{T}_i$}
            \STATE Initialize local policy/value nets: $\theta_k^i \leftarrow \theta$, $\phi_k^i \leftarrow \phi$
            \STATE Initialize replay buffer $\mathcal{D}_i \leftarrow \emptyset$
            \FOR{each sampled episode in $\mathcal{T}_i$}
                \FOR{$t=1$ to $T$}
                    \STATE Encode state $s_t$ via GAT: $\mathbf{e}_t \leftarrow \text{GAT}(\mathbf{G}, \mathbf{f}_{1:t})$
                    \STATE Sample $a_t \sim \pi_{\theta_k^i}(\cdot | s_t)$
                    \STATE Execute $a_t$, observe $r_t$ (Eq.~\eqref{eq:reward_design}), $s_{t+1}$
                    \STATE Store $(s_t, a_t, r_t, s_{t+1})$ in $\mathcal{D}_i$
                \ENDFOR
            \ENDFOR
            \FOR{$j=1$ to $m$}
                \STATE Sample mini-batch from $\mathcal{D}_i$
                \STATE Compute PPO-clip loss $L^{\text{CLIP}}$ (Eq.~\eqref{eq:ppo_clip}) + value loss $L^V$
                \STATE Update $\theta_k^i \leftarrow \theta_k^i + \alpha \nabla L$, $\phi_k^i \leftarrow \phi_k^i + \alpha \nabla L^V$
            \ENDFOR
            \STATE Compute local update delta $\Delta_k^i = \hat{\theta}_k^i - \theta$
        \ENDFOR
        \STATE Upload $\{\Delta_k^i\}$ to aggregation server
    \ENDFOR
    \STATE Aggregate: $\theta \leftarrow \theta + \beta \cdot \frac{1}{KB} \sum_{k,i} \Delta_k^i$ \quad (FedAvg + first-order MAML)
\ENDFOR
\RETURN Meta-policy parameters $\theta$
\end{algorithmic}
\end{algorithm}

At inference, each MEC server performs $m'$ ($m' \ll m$) local PPO updates on newly observed $\mathcal{T}_i$ starting from the federated meta-parameters $\theta$, yielding a customized, dependency-aware offloading policy with minimal overhead.

\section{Experiment}\label{sec:performance_evaluation}

\subsection{Default Setting}

\begin{table}[htbp] \centering \caption{Default Parameter Settings of the Simulated VEC System} \label{tab:default_parameters} \resizebox{\columnwidth}{!}{ \begin{tabular}{lll} \toprule \textbf{Category} & \textbf{Parameter} & \textbf{Default Setting / Range} \\ \midrule \multicolumn{3}{c}{\textbf{Wireless Communication}} \\ \midrule Subchannels per MEC server & Number of subchannels & 4 \\ Subchannel bandwidth & $B$ & $[3,6]$ MHz \\ Channel noise power & $N_0$ & $10^{-5}$ mW \\ Channel gain & $h$ & $[1,3]$ \\ \midrule \multicolumn{3}{c}{\textbf{MEC Server Configuration}} \\ \midrule Processors per MEC & Number of processors & 3 \\ Processor computing rate & $f_m$ & $[2,3]$ G cycles/s \\ \midrule \multicolumn{3}{c}{\textbf{Vehicle Configuration}} \\ \midrule Vehicle computing rate & $f_v$ & $[1,2]$ G cycles/s \\ Transmission power & $P_v$ & $[100,200]$ mW \\ \midrule \multicolumn{3}{c}{\textbf{Task DAG Parameters}} \\ \midrule $n$ & Number of subtasks & 20 \\ $density$ & DAG density & 0.8 \\ $fat$ & DAG shape factor & 0.5 \\ $ccr$ & Computation-to-communication ratio & 0.5 \\ Subtask size & $D$ & $[200, 400]$ KB \\ CPU cycles & $C$ & $[50,60]$ M cycles \\ \bottomrule \end{tabular} } \end{table}

We evaluate the proposed Federated Meta-DRL with GAT-Seq2Seq (FedMAGS) framework in a custom-built simulator that implements the system model in Section~III and the delay formulation in Section~\ref{sec:system_model}.

Heterogeneous vehicular applications are generated using the synthetic task graph generator DAGGEN~\cite{Charles2013}. DAGGEN allows flexible configuration of key DAG parameters, including \textit{n} (number of subtasks), \textit{density} (inter-level dependency connectivity), \textit{fat} (maximum parallelism width), and \textit{ccr} (computation-to-communication ratio). These parameters enable systematic control of structural complexity and workload heterogeneity.

The default system settings are summarized in Table~\ref{tab:default_parameters}. For comparison, we implement three baselines: 1) \textbf{SMRL-MTO}~\cite{Dai2024meta}, a meta-RL approach with Seq2Seq modeling; 2) \textbf{PPO}, where each MEC server independently trains a local policy; and 3) \textbf{DQN}, a value-based offloading scheme with local replay memory. All methods use comparable network configurations to ensure fair evaluation.

\subsection{Pretraining Performance}

\begin{figure}[htp!]
    \centering
    \includegraphics[width=1\linewidth]{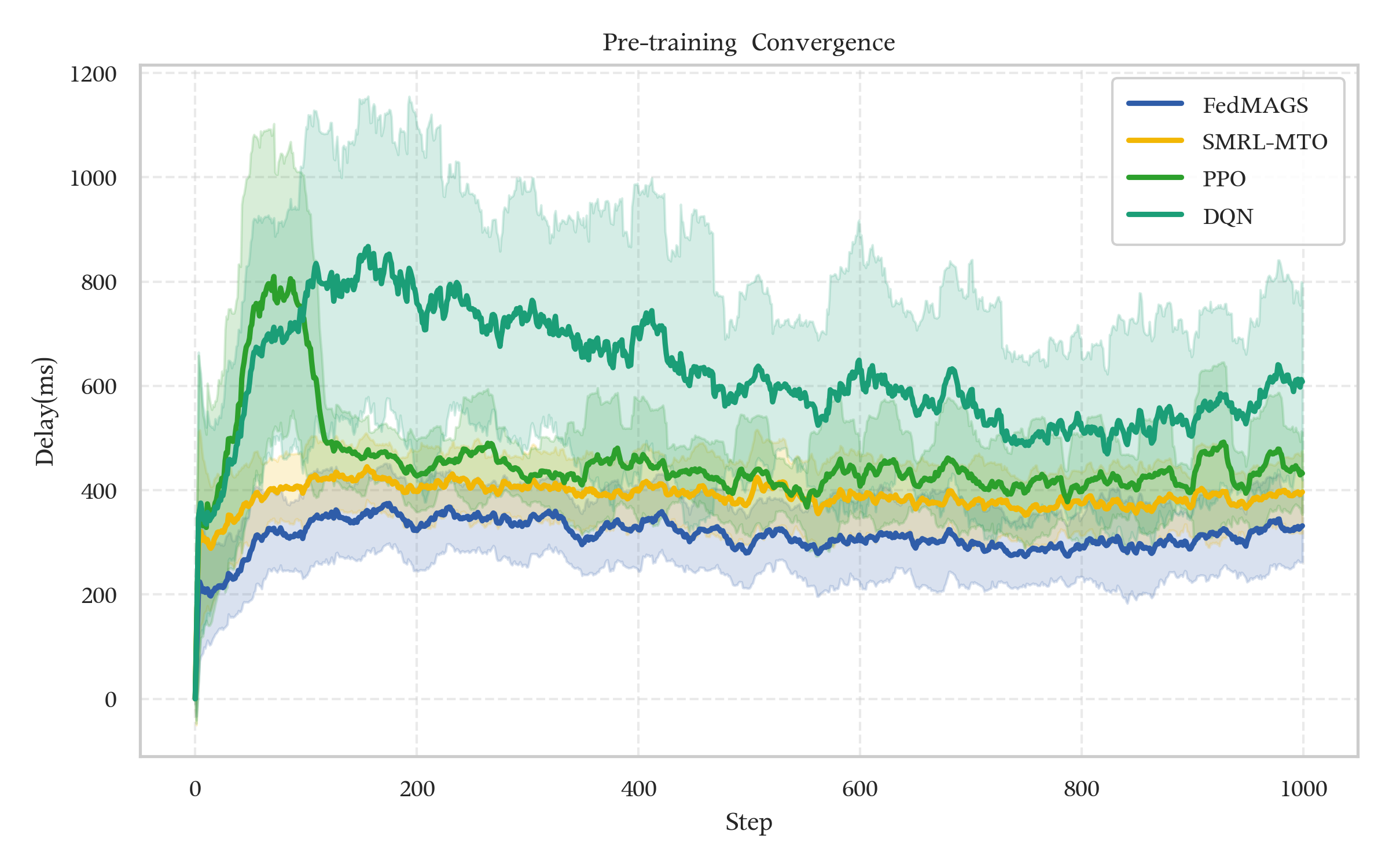}
    \caption{Pretraining Performance of Four Algorithms.}
    \label{fig:pretraining_performance}
\end{figure}

Fig.~\ref{fig:pretraining_performance} compares the pretraining convergence of FedMAGS, SMRL-MTO, PPO, and DQN under the default settings. FedMAGS achieves the fastest convergence and the lowest steady-state AET. Benefiting from federated meta-initialization, the model quickly captures shared knowledge across MEC servers, significantly accelerating learning compared with standalone PPO and DQN. SMRL-MTO improves over independent DRL methods but remains inferior to FedMAGS due to the lack of federated collaboration.

\subsection{Adaptation Performance under Varying Subtask Numbers}

\begin{figure*}[htp!]
\centering
\includegraphics[width=0.75\textwidth]{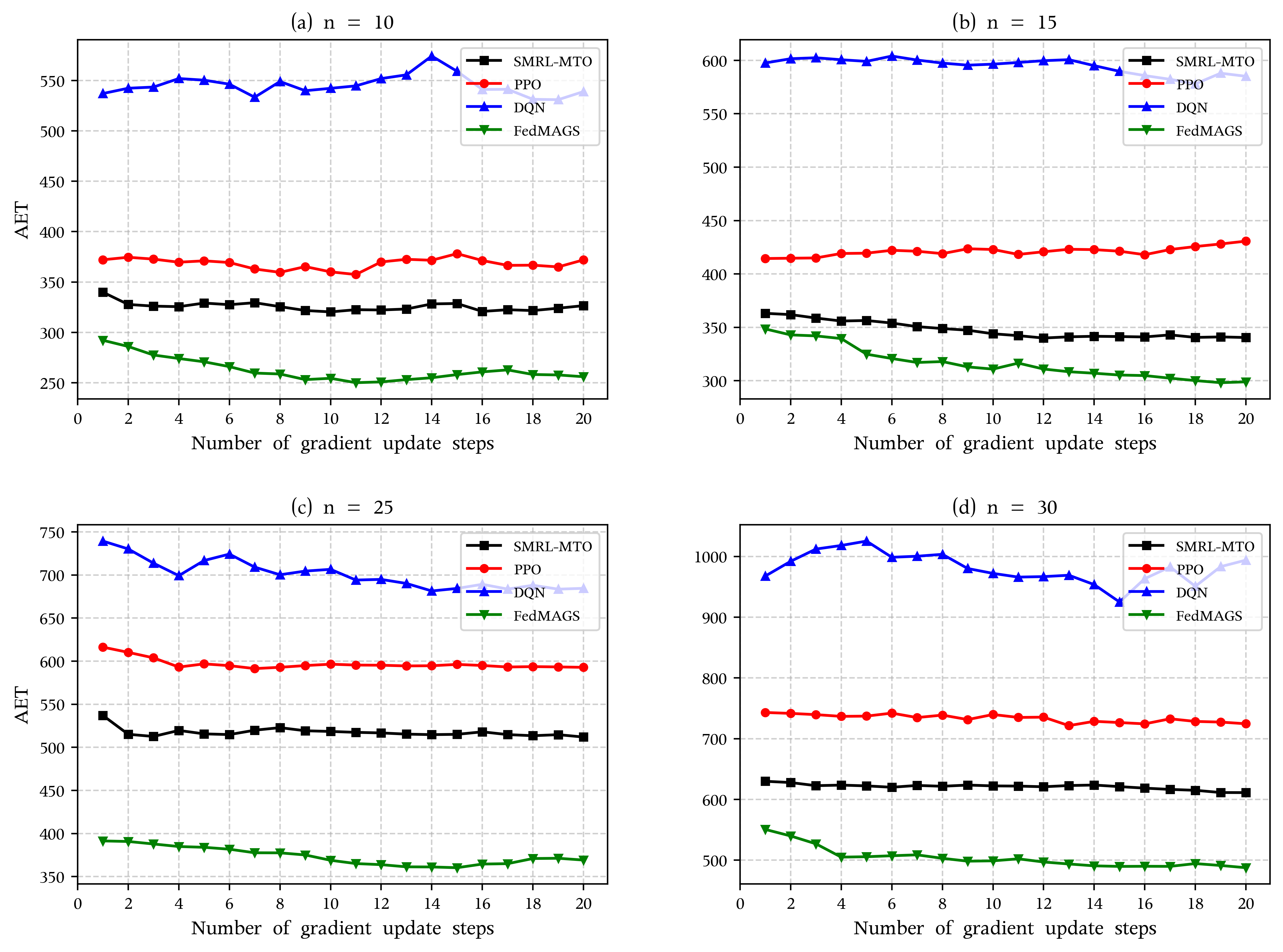}
\caption{Adaptation Performance under Varying Subtask Numbers.}
\label{fig:subtask_performance}
\end{figure*}

We further evaluate adaptation capability by varying the number of subtasks $n \in \{10,15,25,30\}$. As shown in Fig.~\ref{fig:subtask_performance}, FedMAGS consistently achieves the lowest AET and converges within a few gradient steps across all task scales. As $n$ increases, all methods experience higher AET due to increased dependency complexity; however, FedMAGS maintains stable performance and the smallest degradation. In contrast, PPO and SMRL-MTO converge to higher delays, while DQN shows instability for large DAGs. These results demonstrate the superior scalability and fast adaptation capability of FedMAGS.

\subsection{Adaptation Performance under Varying DAG Topologies}
\begin{figure*}[htp!]
\centering
\includegraphics[width=0.75\textwidth]{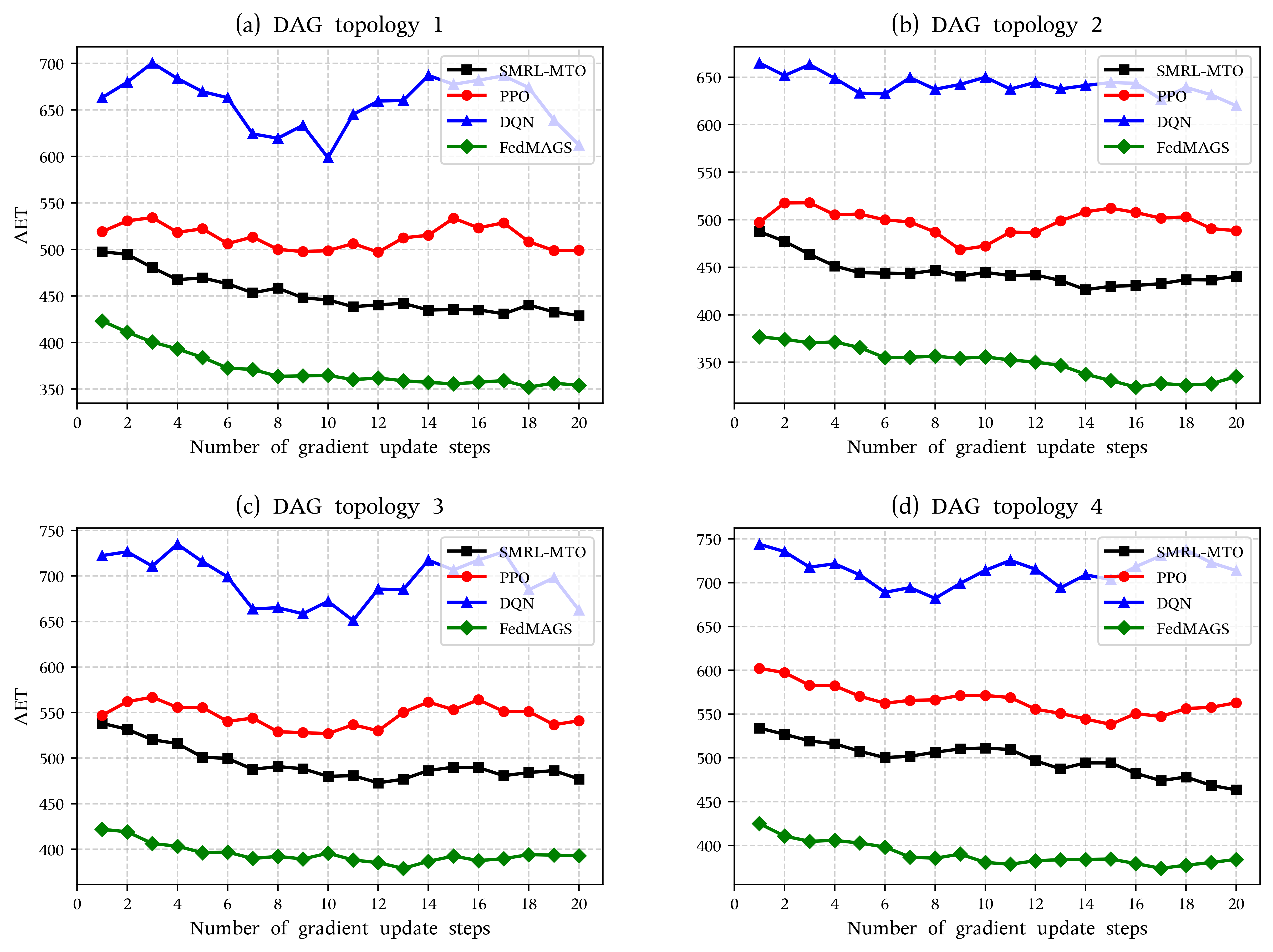}
\caption{Adaptation Performance under Varying DAG Topologies.}
\label{fig:dag_performance}
\end{figure*}

Fig.~\ref{fig:dag_performance} evaluates adaptation performance under four DAG topologies with fixed subtask number $n=20$. The four structures are generated by crossing \textit{density} $\in \{0.7, 0.9\}$ and \textit{fat} $\in \{0.4, 0.6\}$, corresponding to:
Topology1 $(0.7,0.4)$,
Topology2 $(0.7,0.6)$,
Topology3 $(0.9,0.4)$, and
Topology4 $(0.9,0.6)$.
Higher density leads to stronger inter-subtask dependency coupling, while smaller fat reduces parallelism, resulting in more constrained and complex scheduling scenarios.

For Topology1 and Topology2 (moderate structural shift), both FedMAGS and SMRL-MTO exhibit decreasing AET, while FedMAGS converges faster and achieves lower steady-state delay. PPO and DQN show strong oscillations without clear convergence.

Under more complex topologies (Topology3 and Topology4), the performance gap becomes more pronounced. FedMAGS maintains stable convergence and the lowest AET across all steps. In contrast, SMRL-MTO converges more slowly and stabilizes at higher delay values, indicating weakened generalization under dense and constrained DAG structures. PPO and DQN remain unstable, with DQN exhibiting the largest fluctuations.

Overall, FedMAGS demonstrates superior robustness and adaptation capability across varying DAG topologies.

\subsection{Privacy}

In the proposed framework, raw vehicular task data, including DAG structures and execution statistics, remain locally stored at each MEC server. Only model parameter updates are transmitted to the aggregation server during federated training. This design prevents exposure of sensitive information such as vehicle trajectories and workload patterns, while still enabling collaborative policy learning across distributed regions. Moreover, by avoiding raw data transmission, the framework significantly reduces backhaul communication overhead. Consequently, FedMAGS achieves improved generalization and performance without sacrificing data privacy.

\section{Conclusion}

This paper studied heterogeneous DAG-based task offloading in vehicular edge computing and proposed a Federated Meta-DRL framework with GAT-Seq2Seq modeling. By integrating dependency-aware representation learning and federated meta-training, the proposed method achieves fast adaptation, improved scalability, and lower execution delay compared with existing baselines. Moreover, the federated design preserves data privacy and avoids raw data transmission overhead. Experimental results validate the effectiveness and robustness of FedMAGS in dynamic and heterogeneous VEC environments.

\bibliographystyle{IEEEtran}
\bibliography{Refs}{}

\end{document}